\pdfoutput=1
\documentclass[letterpaper,11pt]{article}
\usepackage[hyperref]{dht}
\usepackage{times}
\usepackage{latexsym}

\usepackage{url}

\usepackage{amsmath}
\usepackage{multirow}
\usepackage{graphicx}  
\usepackage{amsfonts}
\usepackage{subcaption}
\renewcommand{\vec}[1]{\mathbf{#1}} 

\DeclareMathOperator*{\argmax}{arg\,max}

\usepackage{footnote}
\usepackage{epstopdf}
\usepackage{algorithm} 
\usepackage{algpseudocode} 

\usepackage{makecell}
\usepackage{color}

\usepackage{tikz}
\usepackage{pgf}
\usepackage{tikz-qtree}
\usepackage{flushend}

\usetikzlibrary{arrows,decorations.pathmorphing,backgrounds,positioning,fit,petri,shapes.misc, arrows.meta,shapes.geometric,decorations.markings,calc,shadows.blur,decorations.pathreplacing,quotes,intersections}
\usepackage{CJKutf8}
\usepackage{pgfplots}
\pgfmathdeclarefunction{gauss}{2}{%
	\pgfmathparse{1/(#2*sqrt(2*pi))*exp(-((x-#1)^2)/(2*#2^2))}%
}
\pgfmathdeclarefunction{myuniform}{1}{%
	\pgfmathparse{1/3}%
}
\definecolor{myblue}{RGB}{20,162,212}
\definecolor{myorange}{RGB}{211, 84, 0}
\definecolor{lowblue}{RGB}{102,178,255}
\definecolor{justblue}{RGB}{84, 160, 255}
\definecolor{mypurple}{RGB}{108, 92, 231}
\definecolor{mygray}{RGB}{158, 158, 158}
\definecolor{lowpurple}{RGB}{204,153,255}
\definecolor{lowwhite}{RGB}{255,255,255}
\definecolor{verylowpurple}{RGB}{255,102,102}
\definecolor{embcolor}{RGB}{255,255,255}
\definecolor{myred}{RGB}{231, 76, 60}
\definecolor{mygreen}{RGB}{162, 217, 206} 
\definecolor{fontgrey}{RGB}{44, 62, 80}
\definecolor{lowpurple}{RGB}{210, 180, 222}
\definecolor{mypumpkin}{RGB}{229, 152, 102}
\definecolor{lowgreen}{RGB}{171, 235, 198}
\definecolor{lowgreen2}{RGB}{186, 220, 88}
\definecolor{lowred}{RGB}{245, 183, 177}
\definecolor{lowyellow}{RGB}{241, 196, 15}
\definecolor{mypink}{RGB}{255, 118, 117}

\tikzset{
	old inner xsep/.estore in=\oldinnerxsep,
	old inner ysep/.estore in=\oldinnerysep,
	double circle/.style 2 args={
		circle,
		old inner xsep=\pgfkeysvalueof{/pgf/inner xsep},
		old inner ysep=\pgfkeysvalueof{/pgf/inner ysep},
		/pgf/inner xsep=\oldinnerxsep+#1,
		/pgf/inner ysep=\oldinnerysep+#1,
		alias=sourcenode,
		append after command={
			let     \p1 = (sourcenode.center),
			\p2 = (sourcenode.east),
			\n1 = {\x2-\x1-#1-0.5*\pgflinewidth}
			in
			node [inner sep=0pt, draw, circle, minimum width=2*\n1,at=(\p1),#2, line width=1.3pt] {}
		}
	},
	double circle/.default={2pt}{blue}
}

\tikzset{middlefactor/.style={decoration={
			markings,
			mark= at position #1 with {\pnode[]{}} 
		},postaction={decorate}},
	middlefactor/.default=0.5
}

\def\pnode [#1]#2{
	\node[regular polygon,regular polygon sides=4, minimum size=1pt,fill=black,#1, inner sep = 2.4pt] (#2) {};
}

\newcommand{\squishlist}{
	\begin{list}{$\bullet$}
		{ \setlength{\itemsep}{0pt}
			\setlength{\parsep}{3pt}
			\setlength{\topsep}{3pt}
			\setlength{\partopsep}{0pt}
			\setlength{\leftmargin}{1.5em}
			\setlength{\labelwidth}{1em}
			\setlength{\labelsep}{0.5em} } }

	\newcounter{Lcount}
	\newcommand{\squishlisttwo}{
		\begin{list}{\arabic{Lcount}. }
			{ \usecounter{Lcount}
				\setlength{\itemsep}{0pt}
				\setlength{\parsep}{0pt}
				\setlength{\topsep}{0pt}
				\setlength{\partopsep}{0pt}
				\setlength{\leftmargin}{2em}
				\setlength{\labelwidth}{1.5em}
				\setlength{\labelsep}{0.5em} } }
		
		\newcommand{\squishend}{
	\end{list} }

\aclfinalcopy 

\title{Dependency-based Hybrid Trees for Semantic Parsing}

\author{Zhanming Jie \and Wei Lu\\
  Singapore University of Technology and Design \\
  8 Somapah Road, Singapore, 487372 \\
  {\tt zhanming\_jie@mymail.sutd.edu.sg}, {\tt luwei@sutd.edu.sg} \\
}

\date{}

\begin{document}
\maketitle
\begin{abstract}
  We propose a novel \textit{dependency-based hybrid tree} model for semantic parsing, which converts natural language utterance into machine interpretable meaning representations. 
  Unlike previous state-of-the-art models, the semantic information is interpreted as the latent dependency between the natural language words in our joint representation. 
  Such dependency information can capture the interactions between the semantics and natural language words. 
  We integrate a neural component into our model and propose an efficient dynamic-programming algorithm to perform tractable inference. 
  Through extensive experiments on the standard multilingual GeoQuery dataset with eight languages, we demonstrate that our proposed approach is able to achieve state-of-the-art performance across several languages. Analysis also justifies the effectiveness of using our new dependency-based representation.\footnote{We make our system and code available at \url{http://statnlp.org/research/sp}.}

\end{abstract}

\section{Introduction}
\label{sec:intro}

\begin{figure}[t!]
	\centering
	\scalebox{0.68}{
		\begin{tikzpicture}[node distance=2.0mm and 2.5mm, >=Stealth, 
		semantic/.style={draw=none, minimum height=5mm, rectangle},
		word/.style={draw=none, minimum height=5mm, rectangle},
		olabel/.style={draw=none, circle, minimum height=9mm, minimum width=9mm,line width=1pt, inner sep=2pt, fill=lowblue, text=fontgrey, label={center:\textsc{o}}},
		bperlabel/.style={draw=none, circle, minimum height=9mm, minimum width=9mm,line width=1pt, inner sep=2pt, fill=lowblue, text=black, label={center:\textsc{per}}},
		borglabel/.style={draw=none, circle, minimum height=9mm, minimum width=9mm,line width=1pt, inner sep=2pt, fill=lowblue, text=black, label={center:\textsc{org}}},
		bgpelabel/.style={draw=none, circle, minimum height=9mm, minimum width=9mm,line width=1pt, inner sep=2pt, fill=lowblue, text=black, label={center:\textsc{Misc}}},
		nnlabel/.style={draw=none, circle, minimum height=9mm, minimum width=9mm,line width=1pt, inner sep=2pt, fill=mypumpkin, text=black, label={center:\textsc{Gpe}}},
		invis/.style={draw=none, circle, minimum height=9mm, minimum width=9mm,line width=1pt, inner sep=2pt, fill=none, text=fontgrey},
		chainLine/.style={line width=1pt,-, color=fontgrey}	
		]
		
		\node[semantic](s1) [] {NL: What rivers do not run through Tennessee ?};
		\node[semantic](mr1) [below= of s1, yshift=2mm] {MR: $answer(exclude(river(all), traverse(stateid('tn'))))$}; 
		\node[draw=black, minimum height=47mm, minimum width=100mm, rectangle, line width=0.8pt, rounded corners] (box) [yshift=-27.5mm] {};
		
		\node[word](m1) [below left = of s1, yshift = -4mm, xshift=15mm] {$m_1$};
		\node[word](m2) [below=of m1, yshift=-2.5mm] {$m_2$};
		\node[word](m3) [below left=of m2, yshift=-1mm,xshift=-2.5mm] {$m_3$};
		\node[word](m4) [below right=of m2, yshift=-1mm, xshift=2.5mm] {$m_4$};
		\node[word](m5) [below =of m4, yshift=-2.5mm] {$m_5$};
		\node[word](m6) [below =of m5, yshift=-2.5mm] {$m_6$};
		
		\node[word](m1r)[right= of m1, xshift=12mm, yshift=-7mm] {$m_1$: Q{\small{UERY}} : $answer$ (R{\small{IVER}})};
		\node[word](m2r)[below= of m1r, yshift=4mm, xshift=5.5mm] {$m_2$: R{\small{IVER}} : $exclude$ (R{\small{IVER}}, R{\small{IVER}})};
		\node[word](m3r)[below= of m2r, yshift=4mm, xshift=-11.2mm] {$m_3$: R{\small{IVER}} : $river$ (all)};
		\node[word](m4r)[below= of m3r, yshift=4mm, xshift=5.9mm] {$m_4$: R{\small{IVER}} : $traverse$ (S{\small{TATE}})};
		\node[word](m5r)[below= of m4r, yshift=4mm, xshift=3.3mm] {$m_5$: S{\small{TATE}} : $stateid$ (S{\small{TATE}}N{\small{AME}})};
		\node[word](m6r)[below= of m5r, yshift=4mm, xshift=-1.8mm] {$m_6$: S{\small{TATE}}N{\small{AME}} : ($'tennessee'$)};
		
		\draw [line width=0.8pt,->, color=fontgrey]  (m1) to [] node[]{} (m2);
		\draw [line width=0.8pt,->, color=fontgrey]  (m2) to [] node[]{} (m3);
		\draw [line width=0.8pt,->, color=fontgrey]  (m2) to [] node[]{} (m4);
		\draw [line width=0.8pt,->, color=fontgrey]  (m4) to [] node[]{} (m5);
		\draw [line width=0.8pt,->, color=fontgrey]  (m5) to [] node[]{} (m6);
		
		
		\node[word](w0) [below = of s1, xshift = -48mm, yshift=-62.5mm] {root};
		\node[word](w1) [right = of w0] {{\em What}};
		\node[invis](rootop) [above = of w0, yshift=8mm] {};
		\node[word](w2) [right = of w1] {{\em rivers}};
		\node[word](w3) [right = of w2] {{\em do}};
		\node[word](w4) [right = of w3] {{\em not}};
		\node[word](w5) [right = of w4] {{\em run}};
		\node[word](w6) [right = of w5] {{\em through}};
		\node[word](w7) [right = of w6] {{\em Tennessee}};
		\node[word](w8) [right = of w7] {{\em ?}};
		
		\draw [line width=0.8pt,->, color=fontgrey]  (w0) to [out=60,in=120, looseness=1] node[above, yshift=0mm, color=black]{$m_1$} (w1);
		\draw [line width=0.8pt,->, color=fontgrey]  (w1) to [out=60,in=120, looseness=1] node[above, color=black]{$m_2$} (w4);
		\draw [line width=0.8pt,->, color=fontgrey]  (w4) to [out=130,in=60, looseness=1] node[above, yshift=-1mm, color=black]{$m_3$} (w2);
		\draw [line width=0.8pt,->, color=fontgrey]  (w4) to [out=60,in=120, looseness=1] node[above, yshift=-1mm, color=black]{$m_4$} (w6);
		\draw [line width=0.8pt,->, color=fontgrey]  (w6) to [out=60,in=120, looseness=1] node[above, yshift=-1mm, color=black]{$m_5$} (w7);
		\draw [line width=0.8pt,->, color=fontgrey]  (w7) to [out=90,in=30, looseness=4.8] node[above, yshift=-1mm, color=black]{$m_6$} (w7);
		\end{tikzpicture} 
	}
	\caption{Top: natural language (NL) sentence; middle: meaning representation (MR); bottom:  dependency-based hybrid tree representation.}
	\label{fig:example}
\end{figure}
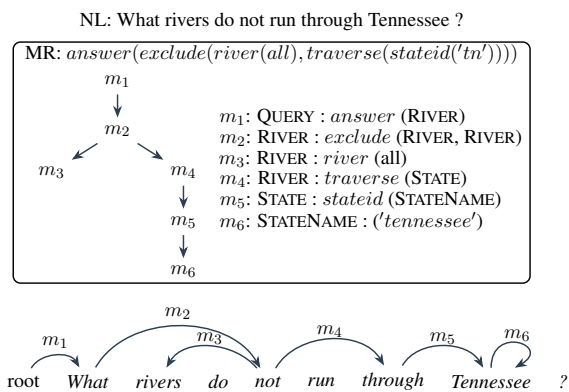

Semantic parsing is a fundamental task within the field of natural language processing (NLP).
Consider a natural language (NL) sentence and its corresponding meaning representation (MR) as illustrated in Figure \ref{fig:example}.
Semantic parsing aims to transform the natural language sentences into machine interpretable meaning representations automatically. 
The task has been popular for decades and keeps receiving significant attention from the NLP community. 
Various systems~\cite{zelle1996learning,kate2005learning,zettlemoyer2005learning,liang11learning} were proposed over the years to deal with different types of semantic representations.
Such models include structure-based models~\cite{wong2006learning,lu2008generative,kwiatkowski2010inducing,jones2012semantic}  and neural network based models~\cite{dong2016language,cheng2017learning}.

Following various previous research efforts \cite{wong2006learning,lu2008generative,jones2012semantic},  in this work, we adopt a popular class of semantic formalism -- logical forms that can be equivalently represented as tree structures.
The  tree representation of an example MR is shown in the middle of Figure \ref{fig:example}.
One challenge associated with building a semantic parser is that the exact correspondence between the words and atomic semantic units are not explicitly given during the training phase.
The key to the building of a successful semantic parsing model lies in the identification of a good {\em joint} latent representation of both the sentence and its corresponding semantics.
Example joint representations proposed in the literature include a chart used in phrase-based translation \cite{wong2006learning}, a constituency tree-like representation known as {\em hybrid tree} \cite{lu2008generative}, and a CCG-based derivation tree \cite{kwiatkowski2010inducing}.

Previous research efforts have shown the effectiveness of using dependency structures to extract semantic representations~\cite{debusmann2004relational,cimiano2009flexible,bedaride2011deep,stanovsky2016getting}. 
Recently, \citet{reddy2016transforming,reddy2017universal} proposed a model to construct logical representations from sentences that are parsed into dependency structures.
Their work demonstrates the connection between the dependency structures of a sentence and its underlying semantics.
Although their setup and objectives are different from ours where externally trained dependency parsers are assumed available and their system was trained to use the semantics for a specific down-stream task, the success of their work motivates us to propose a novel joint  representation that can explicitly capture dependency structures among words for the semantic parsing task.

In this work, we propose a new joint representation for both semantics and words, presenting a new model for semantic parsing. 
Our main contributions can be summarized as follows: 
\squishlist
	\item We present a novel {\em dependency-based hybrid tree} representation that captures both words and semantics in a joint manner.
	Such a dependency tree reveals semantic dependencies between words which are easily interpretable. 
	
	\item We show that exact dynamic programming algorithms for inference can be designed on top of our new representation.  We further show that the model can be integrated with neural networks for improved effectiveness.
	\item Extensive experiments conducted on the standard multilingual GeoQuery dataset 
show that our model outperforms the state-of-the-art models on 7 out of 8 languages. Further analysis confirms the effectiveness of our  dependency-based representation.
\squishend 

To the best of our knowledge, this is the first work that models the semantics as latent dependencies between   words for semantic parsing. 

{\color{red}

}

\section{Related Work}
The literature on semantic parsing has focused on various types of semantic formalisms. 
The $\lambda$-calculus expressions~\cite{zettlemoyer2005learning} have been popular and widely used in semantic parsing tasks over recent years~\cite{dong2016language,gardner2017open,reddy2016transforming,reddy2017universal,susanto2017neural,cheng2017learning}. 
Dependency-based compositional semantics (DCS)\footnote{Unlike ours, their work captures dependencies between  semantic units but not natural language words.} was introduced by \citet{liang11learning}, whose extension, $\lambda$-DCS, was later proposed by \citet{liang2013lambda}. 
Various models~\cite{berant2013semantic,wang2015building,jia-liang:2016:P16-1}  on semantic parsing with the $\lambda$-DCS formalism were proposed. 
In this work, we focus on the tree-structured semantic formalism which has been examined by various research efforts~\cite{wong2006learning,kate2006using,lu2008generative,kwiatkowski2010inducing,jones2012semantic,lu2014semantic,yan2018learn}.

\citet{wong2006learning} proposed the \textsc{Wasp} semantic parser 
that regards the task as a phrase-based machine translation problem.
\citet{lu2008generative} proposed a generative process to generate natural language words and semantic units in a joint model. 
The resulting representation is called \textit{hybrid tree} where both natural language words and semantics are encoded into a joint representation. 
The \textsc{UBL}-s~\cite{kwiatkowski2010inducing} parser applied the CCG grammar~\cite{steedman1996surface} to model the joint representation of both semantic units and contiguous word sequences which do not overlap with one another. 
\citet{jones2012semantic} applied a generative process with Bayesian tree transducer and their model also simultaneously generates the meaning representations and natural language words. 
\citet{lu2014semantic,lu2015constrained} proposed a discriminative version of the hybrid tree model of \cite{lu2008generative} where richer features can be captured. 
\citet{dong2016language} proposed a sequence-to-tree model using recurrent neural networks where the decoder can branch out to produce tree structures. 
\citet{susanto2017semantic} augmented the discriminative {hybrid tree} model with multilayer perceptron and achieved state-of-the-art performance. 

\begin{figure*}[t!]
	\centering
	\scalebox{0.75}{
		\begin{tikzpicture}[node distance=2.0mm and 2.5mm, >=Stealth, 
		semantic/.style={draw=none, minimum height=5mm, rectangle},
		word/.style={draw=none, minimum height=5mm, rectangle},
		olabel/.style={draw=none, circle, minimum height=9mm, minimum width=9mm,line width=1pt, inner sep=2pt, fill=lowblue, text=fontgrey, label={center:\textsc{o}}},
		bperlabel/.style={draw=none, circle, minimum height=9mm, minimum width=9mm,line width=1pt, inner sep=2pt, fill=lowblue, text=black, label={center:\textsc{per}}},
		borglabel/.style={draw=none, circle, minimum height=9mm, minimum width=9mm,line width=1pt, inner sep=2pt, fill=lowblue, text=black, label={center:\textsc{org}}},
		bgpelabel/.style={draw=none, circle, minimum height=9mm, minimum width=9mm,line width=1pt, inner sep=2pt, fill=lowblue, text=black, label={center:\textsc{Misc}}},
		nnlabel/.style={draw=none, circle, minimum height=9mm, minimum width=9mm,line width=1pt, inner sep=2pt, fill=mypumpkin, text=black, label={center:\textsc{Gpe}}},
		invis/.style={draw=none, circle, minimum height=9mm, minimum width=9mm,line width=1pt, inner sep=2pt, fill=none, text=fontgrey},
		chainLine/.style={line width=0.8pt,->, color=fontgrey}	
		]
		\node[semantic](sent) [] {Sentence: {\em What rivers do not run through Tennessee ?}}; 
		
		\node[word](htname) [below left = of sent, xshift=0mm] {{\em Relaxed Hybrid Tree}};
		
		\node[word](hm1) [below = of htname, xshift=0mm] {$m_1$};
		\node[word](hw1) [below left = of hm1, xshift=-20mm, yshift=-3mm] {{\em What}};
		\node[word](hw8) [below right = of hm1, xshift=35mm, yshift=-3mm] {{\em ?}};
		\node[word](hm2) [below = of hm1, xshift=0mm, yshift=-3mm] {$m_2$};
		
		\node[word](hm3) [below left= of hm2, xshift=-8mm, yshift=-3mm] {$m_3$};
		\node[word](hm4) [below right= of hm2, xshift=14mm, yshift=-3mm] {$m_4$};
		\node[word](hw34) [below= of hm2, xshift=0mm, yshift=-3mm] {{\em do not}};
		
		\node[word](hw2) [below= of hm3, xshift=0mm, yshift=-3mm] {{\em rivers}};
		
		\node[word](hm5) [below right= of hm4, xshift=5mm, yshift=-3mm] {$m_5$};
		\node[word](hw56) [below left= of hm4, xshift=2mm, yshift=-3mm] {{\em run through}};
		\node[word](hm6) [below= of hm5, xshift=0mm, yshift=-3mm] {$m_6$};
		\node[word](hw7) [below= of hm6, xshift=0mm, yshift=-3mm] {{\em Tennessee}};
		
		\draw [chainLine] (hm1) to [] node[] {} (hw1);
		\draw [chainLine] (hm1) to [] node[] {} (hm2);
		\draw [chainLine] (hm1) to [] node[] {} (hw8);
		
		\draw [chainLine] (hm2) to [] node[] {} (hm3);
		\draw [chainLine] (hm2) to [] node[] {} (hw34);
		\draw [chainLine] (hm2) to [] node[] {} (hm4);
		
		\draw [chainLine] (hm3) to [] node[] {} (hw2);
		\draw [chainLine] (hm4) to [] node[] {} (hw56);
		\draw [chainLine] (hm4) to [] node[] {} (hm5);
		
		\draw [chainLine] (hm5) to [] node[] {} (hm6);
		\draw [chainLine] (hm6) to [] node[] {} (hw7);

		\node[word](dhtname) [below right = of sent, xshift=-48mm] {{\em Dependency-based Hybrid Tree}};
		\node[word](root) [below = of dhtname, xshift=-5mm] {\textit{root}};
		\node[word](what) [below left = of root, yshift=-3mm, xshift=-3mm] {\textit{What}};
		\node[word](not) [below right = of what, xshift=15mm, yshift=3mm] {\textit{not}};
		\node[word](rivers) [below left = of not, yshift=-4mm] {\textit{rivers}};
		\node[word](through) [below right = of not, yshift=-4mm] {\textit{through}};
		\node[word](tennessee) [below right = of through, yshift=-4mm, xshift=-2mm] {\textit{Tennessee}};
		
		\node[word](m1)[right= of root, xshift=24mm, yshift=2mm] {$m_1$: Q{\small{UERY}} : $answer$ (R{\small{IVER}})};
		\node[word](m2)[below= of m1, yshift=4mm, xshift=5.5mm] {$m_2$: R{\small{IVER}} : $exclude$ (R{\small{IVER}}, R{\small{IVER}})};
		\node[word](m3)[below= of m2, yshift=4mm, xshift=-11.2mm] {$m_3$: R{\small{IVER}} : $state$ (all)};
		\node[word](m4)[below= of m3, yshift=4mm, xshift=5.9mm] {$m_4$: R{\small{IVER}} : $traverse$ (S{\small{TATE}})};
		\node[word](m5)[below= of m4, yshift=4mm, xshift=3.3mm] {$m_5$: S{\small{TATE}} : $stateid$ (S{\small{TATE}}N{\small{AME}})};
		\node[word](m6)[below= of m5, yshift=4mm, xshift=-1.8mm] {$m_6$: S{\small{TATE}}N{\small{AME}} : ($'tennessee'$)};
		
		
		\draw [chainLine] (root) to [] node[above, color=black,sloped] {\small  $m_1$} (what);
		\draw [chainLine] (what) to [] node[above, color=black,sloped] {\small  $m_2$} (not);
		\draw [chainLine] (not) to [] node[above, color=black,sloped] {\small  $m_3$} (rivers);
		\draw [chainLine] (not) to [] node[above, color=black,sloped] {\small  $m_4$} (through);
		\draw [chainLine] (through) to [] node[above, color=black,sloped] {\small  $m_5$} (tennessee);
		\draw [line width=0.8pt,->, color=fontgrey]  (tennessee) to [out=90,in=-30, looseness=5.2] node[above, yshift=-1mm, color=black, sloped]{$m_6$} (tennessee);
		
		\node[word](wroot) [below = of sent, xshift = 5mm, yshift=-64mm] {{\em root}};
		\node[word](w0) [right = of wroot] {{\em What}};
		\node[word](w1) [right = of w0] {{\em rivers}};
		\node[word](w2) [right = of w1] {{\em do}};
		\node[word](w3) [right = of w2] {{\em not}};
		\node[word](w4) [right = of w3] {{\em run}};
		\node[word](w5) [right = of w4] {{\em through}};
		\node[word](w6) [right = of w5] {{\em Tennessee}};
		\node[word](w7) [right = of w6] {{\em ?}};
		
		\draw [line width=0.8pt,->, color=fontgrey]  (wroot) to [out=60,in=120, looseness=1] node[above, yshift=2mm, color=black]{$m_1$} (w0);
		\draw [line width=0.8pt,->, color=fontgrey]  (w0) to [out=60,in=120, looseness=1] node[above, color=black]{$m_2$} (w3);
		\draw [line width=0.8pt,->, color=fontgrey]  (w3) to [out=130,in=60, looseness=1] node[above, yshift=-1mm, color=black]{$m_3$} (w1);
		\draw [line width=0.8pt,->, color=fontgrey]  (w3) to [out=60,in=120, looseness=1] node[above, yshift=-1mm, color=black]{$m_4$} (w5);
		\draw [line width=0.8pt,->, color=fontgrey]  (w5) to [out=60,in=120, looseness=1] node[above, yshift=-1mm, color=black]{$m_5$} (w6);
		\draw [line width=0.8pt,->, color=fontgrey]  (w6) to [out=90,in=30, looseness=4.8] node[above, yshift=-1mm, color=black]{$m_6$} (w6);
		\end{tikzpicture} 
	}
	\caption{The \textit{relaxed hybrid tree} (left)~\cite{lu2014semantic} and our {\em dependency-based hybrid tree} (right) as well as the flat representation (bottom right)  of the example in Figure \ref{fig:example}.}
	\label{fig:dhtexample}
\end{figure*}
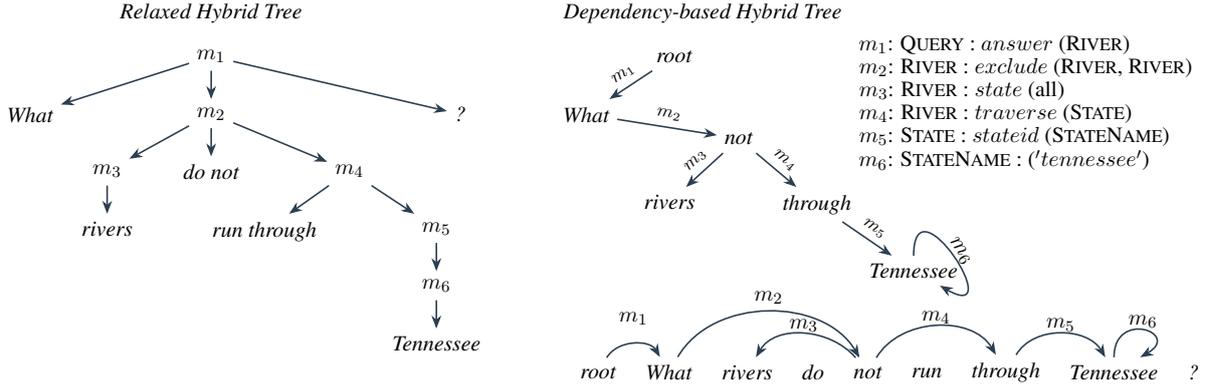

There exists another line of work that applies given syntactic dependency information to semantic parsing. 
\citet{titov2011bayesian} decomposed a syntactic dependency tree into fragments and modeled the semantics as relations between the fragments. 
\citet{poon2013grounded} learned to derive semantic structures based on syntactic dependency trees predicted by the Stanford dependency parser.
\citet{reddy2016transforming,reddy2017universal} proposed a linguistically motivated procedure to transform syntactic dependencies into logical forms. 
Their semantic parsing performance relies on the quality of the syntactic dependencies.
Unlike such efforts, we do not require external syntactic dependencies, but model the semantic units as latent dependencies between natural language words.

\section{Approach}



\subsection{Variable-free Semantics}
\label{sec:semantics}
The variable-free semantic representations in the form of FunQL~\cite{kate2005learning} used by the de-facto GeoQuery dataset~\cite{zelle1996learning} encode semantic compositionality of the logical forms~\cite{cheng2017learning}. 
In the tree-structured semantic representations as illustrated in Figure \ref{fig:example}, each tree node is a semantic unit of the following form:
\begin{equation*}
m_i \equiv \tau_\alpha : p_\alpha(\tau_\beta ^ *)
\end{equation*}
where $m_i$ denotes the complete semantic unit, which consists of semantic type $\tau_\alpha$, function symbol $p_\alpha$ and an argument list of semantic types $\tau_\beta^*$ (here $*$ denotes that there can be 0, 1, or 2 semantic types in the  argument list. This number is known as the {\em arity} of $m_i$). 
Each semantic unit can be regarded as a function that takes in other (partial) semantic representations of certain types as arguments and returns a semantic representation of a specific type.
For example in Figure \ref{fig:example}, the root unit is represented by $m_1$, the type of this unit is \textsc{Query}, the function name is $answer$ and it has a single argument  \textsc{River} which is a semantic type. 
With  recursive function composition, we can obtain a complete MR as shown in Figure \ref{fig:example}.

\subsection{Dependency-based Hybrid Trees}
To jointly encode the tree-structured semantics $\boldsymbol{m}$ and a natural language sentence $\boldsymbol{n}$, we introduce our novel {\em dependency-based hybrid tree}. 
Figure \ref{fig:dhtexample} (right) shows the two equivalent ways of visualizing the {dependency-based hybrid tree} based on the example given in Figure \ref{fig:example}. 
In this example, $\boldsymbol{m}$ is the tree-structured semantics $m_1 (m_2 (m_3, m_4(m_5(m_6))))$ and $\boldsymbol{n}$ is the sentence $\{w_1, w_2, \cdots,w_8\}$\footnote{We also introduce a special token ``{\em root}'' as $w_0$.}. 
Our {dependency-based hybrid tree} $\boldsymbol{t}$ consists of a set of dependencies  between the natural language words, each of which is labeled with a semantic unit. 
Formally, a dependency arc is represented as $(w_p, w_c, m_i)$, where $w_p$ is the {\em parent} of this dependency, $w_c$ is the {\em child}, and $m_i$ is the semantic unit that serves as the {\em label} for the dependency arc. 
A valid dependency-based hybrid tree (with respect to a given semantic representation) allows one to recover the correct semantics from it.
Thus, one constraint is that for any two adjacent dependencies $(w_p, w_c, m_i)$ and $(w_p^\prime, w_c^\prime, m_j)$, where $w_c \equiv w_p^\prime$, $m_i$ must be the parent of $m_j$ in the tree-structured representation $\boldsymbol{m}$.  
For example, in Figure \ref{fig:dhtexample}, the dependencies ({\em not}, {\em through}, $m_4$) and ({\em through}, {\em Tennessee}, $m_5$) satisfy the above condition. 
However, we cannot replace ({\em through}, {\em Tennessee}, $m_5$) with, for example, ({\em through}, {\em Tennessee}, $m_6$), since $m_6$ is not the child of $m_4$. 
Furthermore, the number of children for a word in the dependency tree should be consistent with the arity of the corresponding semantic unit that points to it. 
For example, ``{\em not}'' has 2 children in our {dependency-based hybrid tree} representation because the semantic unit $m_2$ (i.e., R{\small{IVER}} : $exclude$ (R{\small{IVER}}, R{\small{IVER}})) has arity 2. 
Also, ``{\em rivers}'' is the leaf as $m_3$, which points to it, has arity 0.
We will discuss in Section \ref{sec:patterns} on how to derive the set of allowable dependency-based hybrid trees for a given $(\boldsymbol{m},\boldsymbol{n})$ pair.

\begin{table}[t!]
	\centering
	\scalebox{0.65}{
		\begin{tabular}{|c|c|c|}
			\hline 
			Abstract & \multirow{2}{*}{Arity} &\multirow{1}{*}{Dependency} \\
			Semantic Unit & &   Pattern  \\\hline 
			$\mathbf{A}$&0 & $\mathbf{WW}$ \\
			$\mathbf{B}$&1 & $\mathbf{X}$, $\mathbf{WX}$, $\mathbf{XW}$  \\
			$\mathbf{C}$&2 & $\mathbf{XY}$, $\mathbf{YX}$ \\\hline 
		\end{tabular}
	}
	\caption{List of dependency patterns.}
	\label{tab:patterns}
\end{table}

To understand the potential advantages of our new joint representation, we compare it with the {\em relaxed hybrid tree} representation~\cite{lu2014semantic}, which is illustrated on the left of Figure \ref{fig:dhtexample}.
We highlight some similarities and differences between the two representations from the {\em span level} and {\em word level} perspectives.

In a relaxed hybrid tree representation, words and semantic units jointly form a constituency tree-like structure, where the former are  leaves and the latter are internal nodes of such a joint representation.
Such a representation is able to capture alignment between the natural language words and semantics at the span level.\footnote{We refer readers to \cite{lu2014semantic} for more details.}
For example, $m_2$ covers the span from ``{\em rivers}'' to ``{\em Tennessee}'', which allows the interactions between the semantic unit and the span to be captured.
Similarly, in our {dependency-based hybrid tree}, 
such span level word-semantics correspondence can also be captured. 
For example, the arc between ``{\em not}'' and ``{\em through}'' is labeled by the semantic unit $m_4$.
This also allows the interactions between $m_4$ and words within the span from ``{\em not}'' to ``{\em through}'' to be captured.

While both models are able to capture the span-level correspondence between words and semantics,
we can observe that in the relaxed hybrid tree, some words within the span are more directly related to the semantic unit (e.g., ``{\em do not}'' are more related to $m_2$) and some are not.
Specifically, in their representation, the span level information assigned to the parent semantic unit always contains the span level information assigned to all its child semantic units.
This may not always be desirable and may lead to irrelevant features.
In fact, \citet{lu2014semantic} also empirically showed that the span-level features may not always be helpful in their representation.
In contrast, in our {dependency-based hybrid tree}, the span covered by $m_2$ is from ``{\em What}'' to ``{\em not}'', which only consists of the span level information associated with its first child semantic units.
Therefore, our representation is more flexible in capturing the correspondence between words and semantics at the span level, allowing the model to choose the relevant span for features.


Furthermore, our representation can also capture precise interactions between words through  dependency arcs labeled with semantic units.
For example, the semantic unit $m_4$ on the dependency arc from  ``{\em not}'' to ``{\em through}'' in our representation can be used to capture their interactions. 
However, such information could not be straightforwardly captured in a relaxed hybrid tree, which is essentially a constituency tree-like representation.
In the same example, consider the word ``{\em not}'' that bridges  two arcs labeled by $m_2$ and $m_4$. 
Lexical features defined over such arcs can be used to indirectly capture the interactions between  semantic units and guide the tree construction process.
We believe such properties can be beneficial in practice, especially for certain languages. We will examine their significance   in our experiments later.

\begin{figure}[t!]
	\centering
	\scalebox{0.79}{
		\begin{tikzpicture}[node distance=6.0mm and 28mm, >=Stealth, 
		semantic/.style={draw=none, minimum height=5mm, rectangle},
		word/.style={draw=none, minimum height=5mm, rectangle},
		olabel/.style={draw=none, circle, minimum height=9mm, minimum width=9mm,line width=1pt, inner sep=2pt, fill=lowblue, text=fontgrey, label={center:\textsc{o}}},
		bperlabel/.style={draw=none, circle, minimum height=9mm, minimum width=9mm,line width=1pt, inner sep=2pt, fill=lowblue, text=black, label={center:\textsc{per}}},
		borglabel/.style={draw=none, circle, minimum height=9mm, minimum width=9mm,line width=1pt, inner sep=2pt, fill=lowblue, text=black, label={center:\textsc{org}}},
		bgpelabel/.style={draw=none, circle, minimum height=9mm, minimum width=9mm,line width=1pt, inner sep=2pt, fill=lowblue, text=black, label={center:\textsc{Misc}}},
		nnlabel/.style={draw=none, circle, minimum height=9mm, minimum width=9mm,line width=1pt, inner sep=2pt, fill=mypumpkin, text=black, label={center:\textsc{Gpe}}},
		invis/.style={draw=none, circle,line width=0pt, inner sep=0pt, fill=none, text=fontgrey},
		chainLine/.style={line width=0.8pt,->, color=fontgrey}	
		]
		\node[invis](riversparent) {};
		\node[word, below = of riversparent](rivers) {{\em rivers}};
		\draw [chainLine] (riversparent) to [] node[right, color=black] {\small  $m_3$} (rivers);
		\node[word, below left = of rivers, xshift=30mm,yshift=0mm](riversleft) {(...)};
		\node[word, below right = of rivers, xshift=-30mm,yshift=0mm](riversright) {(...)};
		\draw [chainLine, dashed] (rivers) to [] node[right, color=black] {} (riversleft);
		\draw [chainLine, dashed] (rivers) to [] node[right, color=black] {} (riversright);
		
		\node[invis, right =of riversparent, xshift=-5mm](throughparent) {};
		\node[word, below = of throughparent](through) {{\em through}};
		\node[word, below right = of through, xshift=-37mm,yshift=1mm](tn) {{\em Tennessee}};
		\node[word, below left = of through, xshift=32mm,yshift=1mm](tnleft) {(...)};
		\draw [chainLine] (throughparent) to [] node[right, color=black] {\small  $m_4$} (through);
		\draw [chainLine] (through) to [] node[right, color=black] {\small  $m_5$} (tn);
		\draw [chainLine, dashed] (through) to [] node[right, color=black] {} (tnleft);
		
		\node[invis, right =of throughparent](notparent) {};
		\node[word, below = of notparent](not) {{\em not}};
		\node[word, below left = of not, xshift=32mm](rivers2) {{\em rivers}};
		\node[word, below right = of not, xshift=-32mm](through2) {{\em through}};

		\draw [chainLine] (notparent) to [] node[right, color=black] {\small  $m_2$} (not);
		\draw [chainLine] (not) to [] node[left, color=black] {\small  $m_3$} (rivers2);
		\draw [chainLine] (not) to [] node[right, color=black] {\small  $m_4$} (through2);
		
		\node[invis, right =of notparent, xshift=-4mm](tnparent) {};
		\node[word, below = of tnparent](tn2) {{\em Tennessee}};
		
		\draw [chainLine] (tnparent) to [] node[right, color=black] {\small  $m_5$} (tn2);
		\draw [chainLine]  (tn2) to [out=-120,in=-60, looseness=5.8] node[below, yshift=0mm, color=black]{$m_6$} (tn2);
		
		\node[word, below = of rivers, yshift=-4.9mm](a) {$\mathbf{A} \rightarrow \mathbf{W}\mathbf{W}$};
		\node[word, below = of through, yshift=-4mm](a) {$\mathbf{B} \rightarrow \mathbf{W}\mathbf{X}$};
		\node[word, below = of not, yshift=-5mm](a) {$\mathbf{C} \rightarrow \mathbf{X}\mathbf{Y}$};
		\node[word, below = of tn2, yshift=-5mm](a) {$\mathbf{B} \rightarrow \mathbf{X}$};
		\end{tikzpicture}
	}
	\caption{Example dependency  patterns used in the {dependency-based hybrid tree} of Figure \ref{fig:dhtexample}.}
	\label{fig:patternexample}
\end{figure}
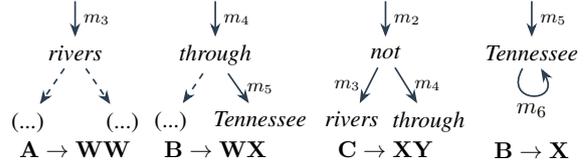

\subsection{Dependency  Patterns}\label{sec:patterns}

To define the set of allowable dependency-based hybrid tree representation so as to allow us to perform exact inference later,
we introduce the {\em dependency  patterns} as shown in Table \ref{tab:patterns}.
We use $\mathbf{A}$, $\mathbf{B}$ or $\mathbf{C}$ to denote the abstract semantic units with arity 0, 1, and 2, respectively. We use $\mathbf{W}$ to denote a contiguous word span, and $\mathbf{X}$ and $\mathbf{Y}$ to denote the first and second child semantic unit, respectively.

We explain these patterns with concrete cases in Figure \ref{fig:patternexample} based on the example in Figure \ref{fig:dhtexample}.
For the first case, the semantic unit $m_3$ has arity 0, the pattern involved is $\mathbf{W}\mathbf{W}$, indicating both the left-hand and right-hand sides of ``{\em rivers}'' (under the dependency arc with semantic unit $m_3$) are just word spans ($\mathbf{W}$, whose length could be zero). 
In the second case, the semantic unit $m_4$ has arity 1, the pattern involved is $\mathbf{WX}$, indicating the left-hand side of ``{\em through}'' (under the arc of semantic unit $m_4$) is a word span  and the right-hand side should be handled by the first child of $m_4$ in the semantic tree, which is $m_5$ in this case.
In the third case, the semantic unit $m_2$ has two arguments, and the pattern involved in the example is $\mathbf{XY}$, meaning the left-hand and right-hand sides should be handled by the first and second child semantic units (i.e., $m_3$ and $m_4$), respectively.\footnote{Analogously, the pattern $\mathbf{YX}$ would mean $m_4$ handles the left-hand side and $m_3$ right-hand side.}
The final case illustrates that we also allow self-loops on our dependency-based hybrid trees, where an arc can be attached to a single word.\footnote{The limitations associated with disallowing such a pattern have been discussed in the previous work of~\cite{lu2015constrained}.} 
To avoid an infinite number of self-loops over a word, we set a maximum depth $c$ to restrict the maximum number of recurrences, which is similar to the method introduced in \cite{lu2015constrained}. 

Based on the dependency patterns, we are able to define the set of all possible allowable {\em dependency-based hybrid tree} representations. Each representation  essentially belongs to a class of {\em projective} dependency trees where semantic units appear on the dependency arcs and (some of the) words are selected as nodes. The semantic tree can be constructed by following the arcs while referring to the dependency patterns involved.

{\color{red}
}

\subsection{Model}


Given the natural language words $\boldsymbol{n}$, our task is to predict  $\boldsymbol{m}$, which is a tree-structured meaning representation, consisting of a set of semantic units as the nodes in the semantic tree. 
We use  $\boldsymbol{t}$ to denote a dependency-based hybrid tree (as shown in Figure \ref{fig:dhtexample}), which jointly encodes both natural language words and the gold meaning representation.  
Let $\mathcal{T}(\boldsymbol{n}, \boldsymbol{m})$ denote all the possible dependency-based hybrid trees that contain the natural language words $\boldsymbol{n}$ and the meaning representation $\boldsymbol{m}$. 
We adopt the widely-used structured prediction model conditional random fields (CRF)~\cite{lafferty2001conditional}. 
The  probability of a possible meaning representation $\boldsymbol{m}$ and dependency-based hybrid tree $\boldsymbol{t}$ for a sentence $\boldsymbol{n}$ is given by:
\begin{equation}
P_\vec{w} (\boldsymbol{m}, \boldsymbol{t} | \boldsymbol{n}) 
= 
\frac{
	e ^{\vec{w}\cdot \vec{f} (\boldsymbol{n}, \boldsymbol{m}, \boldsymbol{t})}
    }
	{
	\sum_{\boldsymbol{m}^\prime, \boldsymbol{t}^\prime \in \mathcal{T}(\boldsymbol{n}, \boldsymbol{m}^\prime )} 
	e^{\vec{w} \cdot \vec{f} (\boldsymbol{n}, \boldsymbol{m}^\prime, \boldsymbol{t}^\prime)}
	}\nonumber
\label{equ:joint}
\end{equation}
where $\vec{f}(\boldsymbol{n}, \boldsymbol{m}, \boldsymbol{t})$ is the feature vector defined over the $(\boldsymbol{n}, \boldsymbol{m}, \boldsymbol{t})$ tuple,
and $\vec{w}$ is the parameter vector.
Since we do not have the knowledge of the ``true'' dependencies during training, $\boldsymbol{t}$ is regarded as a latent-variable in our model. 
We marginalize $\boldsymbol{t}$ in the above equation and the resulting model is a latent-variable CRF~\cite{quattoni2005conditional}:
\begin{equation}
\begin{split}
P_\vec{w} (\boldsymbol{m} | \boldsymbol{n}) 
&= 
\sum_{\boldsymbol{t} \in \mathcal{T} (\boldsymbol{n}, \boldsymbol{m}) }  
P_\vec{w} (\boldsymbol{m}, \boldsymbol{t} | \boldsymbol{n})  \\
& = \frac{
	\sum_{\boldsymbol{t} \in \mathcal{T} (\boldsymbol{n}, \boldsymbol{m}) }	
	e ^{\vec{w} \cdot \vec{f} (\boldsymbol{n}, \boldsymbol{m}, \boldsymbol{t})}
}
{
	\sum_{\boldsymbol{m}^\prime, \boldsymbol{t}^\prime \in \mathcal{T}(\boldsymbol{n}, \boldsymbol{m}^\prime )} 
	e^{\vec{w} \cdot \vec{f} (\boldsymbol{n}, \boldsymbol{m}^\prime, \boldsymbol{t}^\prime)}
}
\end{split}
\label{equ:latent}
\end{equation}

Given a dataset $\mathcal{D}$ of $(\boldsymbol{n}, \boldsymbol{m})$ pairs, our objective is to minimize the negative log-likelihood:\footnote{We ignore the $L_2$ regularization term for brevity.}
\begin{equation}
\begin{split}
\mathcal{L} (\vec{w}) = 
-  \!\!\!\!\!\!
\sum_{(\boldsymbol{n}, \boldsymbol{m}) \in \mathcal{D}}
\!\!\!
\log
\!\!
\sum_{\boldsymbol{t} \in \mathcal{T} (\boldsymbol{n}, \boldsymbol{m}) }  
\!\!
P_\vec{w} (\boldsymbol{m}, \boldsymbol{t} | \boldsymbol{n})
\end{split}
\label{equ:obj}
\end{equation}

The gradient for  model parameter $w_k$ is:
\begin{align}
 \frac{\partial \mathcal{L}(\vec{w})}{\partial w_k}
&=\!\!\!\!\!\!\!
\sum_{(\boldsymbol{n}, \boldsymbol{m}) \in \mathcal{D}}
\sum_{\boldsymbol{m}^\prime, \boldsymbol{t} }
\mathbf{E}_{P_\vec{w}(\vec{m}^\prime, \boldsymbol{t} | \boldsymbol{n}) }
[ f_k (\boldsymbol{n}, \boldsymbol{m}, \boldsymbol{t}) ]\nonumber\\
& - \!\!\!\!\!
\sum_{(\boldsymbol{n}, \boldsymbol{m}) \in \mathcal{D}}
\sum_{\boldsymbol{t} }
\mathbf{E}_{P_\vec{w}(\boldsymbol{t} | \boldsymbol{n}, \boldsymbol{m}) }
[ f_k (\boldsymbol{n}, \boldsymbol{m}, \boldsymbol{t}) ]\nonumber
\end{align}
where $f_k (\boldsymbol{n}, \boldsymbol{m}, \boldsymbol{t})$ represents the number of occurrences of the $k$-th feature. 
With both the objective and gradient above, we can minimize the objective function with standard optimizers, such as L-BFGS~\cite{liu1989limited}  and stochastic gradient descent. 
Calculation of these expectations involves all possible dependency-based hybrid trees.
As there are exponentially many such trees, 
an efficient inference procedure is required.
We will present our efficient algorithm to perform  exact inference for learning and decoding in the next section. 

\subsection{Learning and Decoding}
\label{sec:learninganddecode}

We propose  dynamic-programming algorithms to perform efficient and exact inference, which will be used for calculating the objective and gradients discussed in the previous section.
The algorithms are inspired by the inside-outside style algorithm~\cite{baker1979trainable}, graph-based dependency parsing~\cite{eisner2000bilexical,koo2010efficient,shi2017fast}, and the \textit{relaxed hybrid tree} model~\cite{lu2014semantic,lu2015constrained}. 
As discussed in Section \ref{sec:patterns}, our latent dependency trees are {projective} as in traditional dependency parsing~\cite{eisner1996three,nivre2004deterministic,mcdonald2005online} -- the dependencies are non-crossing with respect to the word order (see bottom of Figure \ref{fig:example}).

The objective function in Equation \ref{equ:obj} can be further decomposed into the following form\footnote{Regularization term is excluded for brevity.}:
\begin{equation}
\begin{split}
\mathcal{L} (\vec{w})
=
-
\sum_{(\boldsymbol{n}, \boldsymbol{m}) \in \mathcal{D}}
\log \sum_{\boldsymbol{t} \in \mathcal{T} (\boldsymbol{n}, \boldsymbol{m}) }	
e ^{\vec{w} \cdot \vec{f} (\boldsymbol{n}, \boldsymbol{m}, \boldsymbol{t})} \\
+
\sum_{(\boldsymbol{n}, \boldsymbol{m}) \in \mathcal{D}}
\!\!\!
\log
\sum_{\boldsymbol{m}^\prime, \boldsymbol{t}^\prime \in \mathcal{T}(\boldsymbol{n}, \boldsymbol{m}^\prime )} 
e^{\vec{w} \cdot \vec{f} (\boldsymbol{n}, \boldsymbol{m}^\prime, \boldsymbol{t}^\prime)}\nonumber
\end{split}
\label{equ:objdecompose}
\end{equation}

We can see the first term is essentially the combined score of all the possible latent structures containing the pair $(\boldsymbol{n}, \boldsymbol{m})$. 
The second term is the combined score for all the possible latent structures containing $\boldsymbol{n}$. 
We show how such scores can be calculated in a factorized manner, based on the fact that we can recursively decompose a dependency-based hybrid tree based on the dependency patterns we introduced.

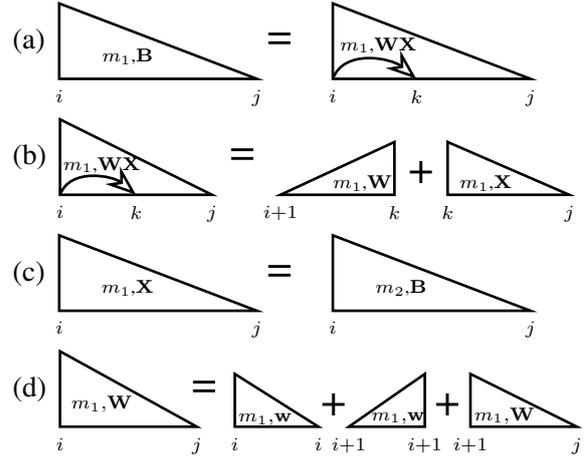
\begin{figure}[t!]
	\centering
	\begin{subfigure}{\columnwidth}
		\begin{tikzpicture}
		\node [scale = 1](a) at (-0.6,0.5) {(a)};
		\draw [line width = 1pt] (2.4,0) node [below] {\scriptsize $j$} -- (-0.2,0) node [below] {\scriptsize $i$} -- (-0.2,1) -- cycle;
		\node [](pa_type1) at (0.7, 0.3) {\scriptsize $m_1$,$ \mathbf{B}$};
		\node [scale = 1.5](equals) at (2.7, 0.5) {=};
		\draw [line width = 1pt] (6.0,0) node [below] {\scriptsize $j$} -- (3.4,0) node [below] {\scriptsize $i$} -- (3.4,1)-- cycle;
		\node [](type) at (4.0, 0.45) {\scriptsize $m_1$,$\mathbf{WX}$};
		\draw [line width=1pt, -{Stealth[length=4mm, open]}] (3.4,0) to [out=70,in=140] node [above] {} (4.5,0);
		\node [](type) at (4.5, -0.24) {\scriptsize $k$};
		\end{tikzpicture}
	\end{subfigure}
	
	\begin{subfigure}{\columnwidth}
		\begin{tikzpicture}
		\node [scale = 1](a) at (-0.6,0.5) {(b)};
		\draw [line width = 1pt] (1.8,0) node [below] {\scriptsize $j$} -- (-0.2,0) node [below] {\scriptsize $i$} -- (-0.2,1) -- cycle;
		\node [](k) at (0.8, -0.24) {\scriptsize $k$};
		\draw [line width=1pt, -{Stealth[length=4mm, open]}] (-0.2,0) to [out=70,in=140] node [above] {} (0.8, 0);
		\node [](type) at (0.35, 0.4) {\scriptsize $m_1$,$\mathbf{WX}$};
		\node [scale = 1.5](equals) at (2.2,0.5) {=};
		\draw [line width = 1pt] (4.2,0) node [below] {\scriptsize $k$} -- (2.7,0) node [below] {\scriptsize $i$$+$$1$}-- (4.2,0.7) -- cycle;
		\node [](pa_type1) at (3.78, 0.15) {\scriptsize $m_1$,$\mathbf{W}$};
		\node [scale = 1.5](equals) at (4.55,0.3) {+};
		\draw [line width = 1pt] (6.5,0) node [below] {\scriptsize $j$} -- (4.9,0) node [below] {\scriptsize $k$} -- (4.9,0.7) -- cycle;
		\node [](pa_type2) at (5.4, 0.15) {\scriptsize $m_1$,$\mathbf{X}$};
		\end{tikzpicture}
	\end{subfigure}
	\begin{subfigure}{\columnwidth}
		\begin{tikzpicture}
		\node [scale = 1](a) at (-0.6,0.5) {(c)};
		\draw [line width = 1pt] (2.4,0) node [below] {\scriptsize $j$} -- (-0.2,0) node [below] {\scriptsize $i$} -- (-0.2,1) -- cycle;
		\node [](pa_type1) at (0.7, 0.3) {\scriptsize $m_1$,$\mathbf{X}$};
		\node [scale = 1.5](equals) at (2.7, 0.5) {=};
		\draw [line width = 1pt] (6.0,0) node [below] {\scriptsize $j$} -- (3.4,0) node [below] {\scriptsize $i$} -- (3.4,1)-- cycle;
		\node [](type) at (4.3, 0.3) {\scriptsize $m_2$,$\mathbf{B}$};
		\end{tikzpicture}
	\end{subfigure}
	\begin{subfigure}{\columnwidth}
		\begin{tikzpicture}
		\node [scale = 1](a) at (-0.6,0.5) {(d)};
		\draw [line width = 1pt] (1.6,0) node [below] {\scriptsize $j$} -- (-0.2,0) node [below] {\scriptsize $i$} -- (-0.2,1) -- cycle;
		\node [](type) at (0.35, 0.3) {\scriptsize $m_1$$,$$\mathbf{W}$};
		\node [scale = 1.5](equals) at (1.7,0.5) {=};
		\draw [line width = 1pt] (3.2,0) node [below] {\scriptsize $i$} -- (2.1,0.0) node [below] {\scriptsize $i$}-- (2.1,0.7) -- cycle;
		\node [](pa_type1) at (2.5, 0.1) {\scriptsize $m_1$$,$$\mathbf{w}$};
		\node [scale = 1.5](equals) at (3.4,0.3) {+};
		\draw [line width = 1pt] (4.6,0) node [below] {\scriptsize $i$$+$$1$} -- (3.6,0) node [below] {\scriptsize $i$$+$$1$} -- (4.6,0.7) -- cycle;
		\node [](pa_type2) at (4.25, 0.12) {\scriptsize $m_1$$,$$\mathbf{w}$};
		\node [scale = 1.5](equals) at (4.9,0.3) {+};
		\draw [line width = 1pt] (6.6,0) node [below] {\scriptsize $j$} -- (5.2,0.0) node [below] {\scriptsize $i$$+$$1$}-- (5.2,0.7) -- cycle;
		\node [](pa_type2) at (5.65, 0.15) {\scriptsize $m_1$$,$$\mathbf{W}$};
		\end{tikzpicture}
	\end{subfigure}
	\caption{The dynamic-programing structures and derivation of our model. The other direction is symmetric. See supplementary material for the complete structures.}
	\label{fig:modelderivation}
\end{figure}

Formally, we introduce two interrelated dynamic-programming structures that are similar to those used in graph-based dependency parsing~\cite{eisner2000bilexical,koo2010efficient,shi2017fast}, namely {\em complete span} and {\em complete arc span}. 
Figure \ref{fig:modelderivation}a shows an example of {\em complete span} (left) and {\em complete arc span} (right). 
The {\em complete span} (over $[i,j]$) consists of a headword (at $i$) and its descendants on one side (they altogether form a subtree), a dependency pattern and a semantic unit. 
The {\em complete arc span} is a span (over $[i,j]$) with a dependency between the headword (at $i$) and the modifier (at $k$). 
We use $C_{i,j,p,m}$ to denote a complete span, where $i$ and $j$ represent the indices of the headword and endpoint, $p$ is the dependency pattern and $m$ is the semantic unit. 
Analogously, we use $A_{i,k,j,p,m}$ to denote a complete arc span where $i$ and $k$ are used to denote the additional dependency from the word at the $i$-th position as headword to the word at the $k$-{th} position as modifier. 

As we can see from the derivation in Figure \ref{fig:modelderivation}, each type of span can be constructed from smaller spans in a bottom-up manner.  
Figure \ref{fig:modelderivation}a shows that a {complete span} is constructed from a {complete arc span} following the dependency patterns in Table \ref{tab:patterns}. 
Figure \ref{fig:modelderivation}b shows a {complete arc span} can be simply constructed from two smaller complete spans based on the dependency pattern.  
In Figure \ref{fig:modelderivation}c and \ref{fig:modelderivation}d, we further show how such two complete spans with pattern $\mathbf{X}$ (or $\mathbf{Y}$) and $\mathbf{W}$ can be constructed.
Figure \ref{fig:modelderivation}c illustrates how to model a transition from one semantic unit to another where the parent is $m_1$ and the child is $m_2$ in the semantic tree. 
If $m_2$ has arity 1, then the pattern is $\mathbf{B}$ following the dependency patterns in Table \ref{tab:patterns}. 
For spans with a single word, we use the lowercase $\mathbf{w}$ as the pattern to indicate this fact, as shown in Figure \ref{fig:modelderivation}d. 
They are the atomic spans used for building larger spans. 
As the {complete span} in Figure \ref{fig:modelderivation}d is associated with pattern $\mathbf{W}$, which means the words within this span are under the semantic unit $m_1$, we can incrementally construct this span with atomic spans.
We illustrate the construction of a complete dependency-based hybrid tree in the supplementary material.

Our final goal during training for a sentence $\boldsymbol{n}=\{w_0, w_1,\cdots, w_N\}$ is to construct all the possible {\em complete spans} that cover the interval $[0, N]$, which can be represented as $C_{0,N,\cdot, \cdot}$. 
Similar to the chart-based dependency parsing algorithms~\cite{eisner1996three,eisner2000bilexical,koo2010efficient}, we can obtain the {\em inside} and {\em outside} scores using our dynamic-programming derivation in Figure \ref{fig:modelderivation} during the inference process,
which can then be used to calculate the objective and feature expectations. 
Since the spans are defined by at most three free indices, the dependency pattern and the semantic unit, our dynamic-programming algorithm requires $\mathcal{O}(N^3M)$ time\footnote{We omit a small constant factor associated with patterns.} where $M$ is the number of semantic units. 
The resulting complexity is the same as the {relaxed hybrid tree} model~\cite{lu2014semantic}. 


%
%
%
%
%
%

During decoding, we can find the optimal (tree-structured) meaning representation $\boldsymbol{m}^*$ for a given  input sentence $\boldsymbol{n}$ by the Viterbi algorithm. 
This step can also be done efficiently with our dynamic-programming approach, where we switch from marginal inference to MAP inference:
\begin{equation*}
\boldsymbol{m}^*, \boldsymbol{t}^* = \argmax_{\boldsymbol{m}, \boldsymbol{t}\in\mathcal{T} (\boldsymbol{n}, \boldsymbol{m})} e ^{\vec{w} \cdot \vec{f} (\boldsymbol{n}, \boldsymbol{m}, \boldsymbol{t})}
\end{equation*}
A similar decoding procedure has been used in previous work~\cite{lu2014semantic,durrett2015neural} with CKY-based parsing algorithm.


\subsection{Features}
\label{sec:features}
As shown in Equation \ref{equ:joint}, the features are defined on the tuple $(\boldsymbol{n}, \boldsymbol{m}, \boldsymbol{t})$. 
With the dynamic-programming procedure, we can define the features over the structures in Figure \ref{fig:dhtexample}. 
Our feature design is inspired by the {hybrid tree} model~\cite{lu2015constrained} and graph-based dependency parsing~\cite{mcdonald2005online}. 
Table \ref{tab:features} shows the feature templates for the example in Figure \ref{fig:dhtexample}. 
Specifically, we define simple unigram features (concatenation of a semantic unit and a word that directly appears under the unit), pattern features (concatenation of the semantic unit and the child pattern) and transition features (concatenation of the parent and child semantic units). 
They form our basic feature set.

\begin{table}[t!]
	\centering
	\scalebox{0.70}{
		\begin{tabular}{|l|c|}
			\hline
			Feature Type & Examples \\
			\hline
			\hline 
			Word &  ``$m_4$ \& {\em run}'', ~~``$m_4$ \& {\em through}'' \\
			Pattern & ``$m_2$ \& $\mathbf{XY}$'', ~~``$m_4$ \& $\mathbf{WX}$'' \\
			Transition & ``$m_2$ \& $m_3$'', ~~``$m_2$ \& $m_4$'' \\
			\hline
			\hline 
			Head word & ``$m_2$ \& {\em What}'', ~~``$m_4$ \& {\em not}'' \\
			Modifier word & ``$m_2$ \& {\em not}'', ~~``$m_4$ \& {\em through}'' \\
			Bag of words & ``$m_4$ \& {\em not}'', ~~``$m_4$ \& {\em run}'', ~~``$m_4$ \& {\em through}'' \\\hline 
		\end{tabular}
	}
	\caption{Features for the example in Figure \ref{fig:dhtexample}. 
	}
	\label{tab:features}
\end{table}

Additionally, with the structured properties of dependencies, we can define dependency-related features~\cite{mcdonald2005online}. 
We use the parent (head) and child (modifier) words of the dependency as features.
We also use the bag-of-words covered under a dependency as features. 
The dependency features are useful in helping improve the performance as we can see in the experiments section. 

\subsection{Neural Component}

Following the approach used in \citet{susanto2017semantic}, we could further incorporate  neural networks into our latent-variable graphical model. 
The integration is analogous to the approaches described in the neural CRF models~\cite{artieres2010neural,durrett2015neural,gormley2015graphical,lample2016neural},
where we use neural networks to learn distributed feature representations within our graphical model.

We employ a neural architecture to calculate the score associated with each dependency arc $(w_p, w_c, m)$ (here $w_p$ and $w_c$ are the parent and child words in the dependency and $m$ is the semantic unit over the arc), where the input to the neural network consists of  words (i.e., $(w_p, w_c)$) associated with this dependency and the neural network will calculate a score for each possible semantic unit, including $m$.
The two words are first mapped to word embeddings  $\vec{e}_p$ and $\vec{e}_c$ (both of dimension $d$). 
Next, we use a bilinear layer\footnote{Empirically, we also tried multilayer perceptron but the bilinear model gives us better results.}~\cite{socher2013recursive,chen2016thorough} to capture the interaction between the parent and the child in a dependency:
\begin{equation*}
r_i = \vec{e}_p^{\mathbf{T}} \mathbf{U}_i \vec{e}_c
\end{equation*}
where $r_i$ represents the score for the $i$-th semantic unit and $\mathbf{U}_i \in \mathbb{R}^{d\times d}$.
The scores are then incorporated into the probability expression in Equation \ref{equ:joint} during learning and decoding.
As a comparison, we also implemented a  variant where our model directly takes in the average embedding of $\mathbf{e}_p$ and $\mathbf{e}_c$ as additional features, without using our neural component. 


\begin{table*}[t!]
	\centering
	\scalebox{0.55}{
		\begin{tabular}{|c|l|cc|cc|cc|cc|cc|cc|cc|cc|}
			\hline 
			\multirow{2}{*}{Type}&\multirow{2}{*}{System/Model} & \multicolumn{2}{c|}{English (en)}& \multicolumn{2}{c|}{Thai (th)}& \multicolumn{2}{c|}{German (de)}& \multicolumn{2}{c|}{Greek (el)}& \multicolumn{2}{c|}{Chinese (zh)}& \multicolumn{2}{c|}{Indonesian (id)}& \multicolumn{2}{c|}{Swedish (sv)}& \multicolumn{2}{c|}{Farsi (fa)} \\
			&& \textit{Acc.} & \textit{F.}& \textit{Acc.} & \textit{F.}& \textit{Acc.} & \textit{F.}& \textit{Acc.} & \textit{F.}& \textit{Acc.} & \textit{F.}& \textit{Acc.} & \textit{F.}& \textit{Acc.} & \textit{F.}& \textit{Acc.} & \textit{F.} \\
			\hline 
			\hline 
			\multirow{5}{*}{Non-Neural}&\textsc{Wasp}  & 71.1 & 77.7 & 71.4 & 75.0 & 65.7 & 74.9 & 70.7 & 78.6 & 48.2 & 51.6 & 74.6 & 79.8 & 63.9 & 71.5 & 46.8 & 54.1 \\
			&\textsc{HybridTree}  & 76.8 & 81.0 & 73.6 & 76.7 & 62.1 & 68.5 & 69.3 & 74.6 & 56.1 & 58.4 & 66.4 & 72.8 & 61.4 & 70.5 & 51.8 & 58.6 \\
			&\textsc{UBL}  & 82.1 & 82.1 & 66.4 &66.4 & 75.0 & 75.0 & 73.6 &73.7 & 63.8 & 63.8 & 73.8 & 73.8 & 78.1 & 78.1 &64.4 & 64.4\\
			&\textsc{TreeTrans}  & 79.3 & 79.3 & 78.2 &78.2 & 74.6 &74.6 & 75.4 &75.4 & - & - &- & -&- & -&- & -\\
			&\textsc{RHT}  & 86.8 & 86.8 & 80.7 & 80.7 &75.7 & 75.7 & 79.3 & 79.3 & 76.1 & 76.1 & 75.0 & 75.0 &  79.3 & 79.3 &  73.9 & 73.9\\\hline 
			\hline 
			\multirow{5}{*}{Neural}&\textsc{Seq2Tree}$\dagger$  & 84.5 & - & 71.9 & - & 70.3 & - & 73.1 & - & 73.3 & - & 80.7 & - & 80.8 & - & 70.5 & - \\
			&\textsc{MSP-Single}$\dagger$ & 83.5 & - & 72.1 & - & 69.3 & - &74.2 & - & 74.9 & - & 79.8 & - & 77.5 & - & 72.2 & - \\
			&\textsc{Neural HT} ($J$=0)&  87.9&87.9  &82.1 &82.1 &75.7&75.7 &81.1 &81.1&76.8&76.8&76.1&76.1&81.1&81.1&75.0&75.0 \\
			&\textsc{Neural HT} ($J$=1)&88.6  &88.6 &84.6 &84.6  &76.8 &76.8 &79.6 &79.6&75.4&75.4&78.6&78.6&82.9&82.9&76.1 &76.1 \\
			&\textsc{Neural HT} ($J$=2)& \textbf{90.0} &\textbf{90.0}  &  82.1&82.1  & 73.9&73.9 &80.7 &80.7&81.1&81.1&{81.8}&{81.8}&83.9&83.9&74.6&74.6\\ \hline\hline
			Non-Neural&(This work) \textsc{DepHT}  & 86.8 & 86.8 & 81.8 & 81.8 & 76.1 & 76.1 & 80.4 & 80.4 & 81.4 & 81.4& 86.8 & 86.8 & 85.4 & 85.4 & 73.9& 73.9    \\
			Non-Neural&(This work) \textsc{DepHT} + embedding & 87.5 & 87.5 &83.9 & 83.9 & 75.0 & 75.0 & 81.1 & 81.1 & 81.4 & 81.4& 87.5 & 87.5 & 87.1 & 87.1 & 73.6 & 73.6    \\
			Neural&(This work) \textsc{DepHT} + NN  & 89.3 & 89.3 & \textbf{86.7} & \textbf{86.7}& {\bf 78.2} & {\bf 78.2} &  {\bf 82.9}& {\bf 82.9}& {\bf 82.9} & {\bf 82.9} & {\bf 88.7} & {\bf 88.7} & {\bf 87.3} & {\bf 87.3} & {\bf 77.9}& {\bf 77.9} \\\hline 
			
		\end{tabular}
	}
	\caption{Performance comparison with state-of-the-art models on GeoQuery dataset. ($\dagger$ represents the system is using lambda-calculus expressions as meaning representations.)}
	\label{tab:nonneuralresults}
\end{table*}

\section{Experiments}

\paragraph{Data and evaluation methodology} 

We conduct experiments on the publicly available variable-free version of the GeoQuery dataset, which has been widely used for semantic parsing~\cite{wong2006learning,lu2008generative,jones2012semantic}. 
The dataset consists of 880 pairs of natural language sentences and the corresponding tree-structured semantic representations. 
This dataset is annotated with eight languages. 
The original annotation of this dataset is English~\cite{zelle1996learning} and \citet{jones2012semantic} annotated the dataset with three more languages: German, Greek and Thai. 
\citet{lu2011probabilistic} released the Chinese annotation and \citet{susanto2017semantic} annotated the corpus with three additional languages: Indonesian, Swedish and Farsi. 
In order to compare with previous work~\cite{jones2012semantic,lu2015constrained}, we follow the standard splits with 600 instances for training and 280 instances for testing. 
To evaluate the performance, we follow the standard evaluation procedure used in various previous works \cite{wong2006learning,lu2008generative,jones2012semantic,lu2015constrained} to construct the Prolog query from the tree-structured semantic representation using a standard and publicly available script. 
The queries are then used to retrieve the answers from the GeoQuery database, and we report  accuracy and $F_1$ scores.

\paragraph{Hyperparameters} 
We set the maximum depth $c$ of the semantic tree to 20, following \citet{lu2015constrained}. 
The $L_2$ regularization coefficient is tuned from 0.01 to 0.05 using 5-fold cross-validation on the training set.
The Polyglot~\cite{polyglot:2013:ACL-CoNLL} multilingual word embeddings\footnote{The embeddings are fixed to avoid overfitting.} (with 64 dimensions) are used for all languages.  
We use L-BFGS~\cite{liu1989limited} to optimize the \textsc{DepHT} model until convergence 
and stochastic gradient descent (SGD) with a learning rate of 0.05 to optimize the neural \textsc{DepHT} model.
We implemented our neural component with the Torch7 library~\cite{collobert2011torch7}. 
Our complete implementation is based on the StatNLP\footnote{https://gitlab.com/sutd\_nlp/statnlp-core} structured prediction framework~\cite{lu2017unified}.

\subsection{Baseline Systems}

We run the released systems of several state-of-the-art semantic parsers, namely the \textsc{Wasp} parser~\cite{wong2006learning}, \textsc{HybridTree} model~\cite{lu2008generative}, \textsc{UBL} system~\cite{kwiatkowski2010inducing},  \textit{relaxed hybrid tree} (\textsc{RHT})~\cite{lu2015constrained}\footnote{\cite{lu2015constrained} is an extension of the original \textit{relaxed hybrid tree}~\cite{lu2014semantic}, which reports improved results.}, the sequence-to-tree (\textsc{Seq2Tree}) model~\cite{dong2016language}, the \textit{neural hybrid tree} (\textsc{Neural HT}) model~\cite{susanto2017semantic}, and the multilingual semantic parser~\cite{susanto2017neural} with single language (\textsc{MSP-Single}) as input. 
The results for \textsc{TreeTrans}~\cite{jones2012semantic} are taken from their paper.

\begin{figure}[t!]
	\centering
	\scalebox{0.55}{
		\begin{tikzpicture}[node distance=2.0mm and 2.5mm, >=Stealth, 
		semantic/.style={draw=none, minimum height=5mm, rectangle},
		word/.style={draw=none, minimum height=5mm, rectangle},
		olabel/.style={draw=none, circle, minimum height=9mm, minimum width=9mm,line width=1pt, inner sep=2pt, fill=lowblue, text=fontgrey, label={center:\textsc{o}}},
		bperlabel/.style={draw=none, circle, minimum height=9mm, minimum width=9mm,line width=1pt, inner sep=2pt, fill=lowblue, text=black, label={center:\textsc{per}}},
		borglabel/.style={draw=none, circle, minimum height=9mm, minimum width=9mm,line width=1pt, inner sep=2pt, fill=lowblue, text=black, label={center:\textsc{org}}},
		bgpelabel/.style={draw=none, circle, minimum height=9mm, minimum width=9mm,line width=1pt, inner sep=2pt, fill=lowblue, text=black, label={center:\textsc{Misc}}},
		nnlabel/.style={draw=none, circle, minimum height=9mm, minimum width=9mm,line width=1pt, inner sep=2pt, fill=mypumpkin, text=black, label={center:\textsc{Gpe}}},
		invis/.style={draw=none, circle, minimum height=9mm, minimum width=9mm,line width=1pt, inner sep=2pt, fill=none, text=fontgrey},
		chainLine/.style={line width=0.8pt,->, color=fontgrey}	
		]
		\node[semantic](sent) [] {Sentence: {\em San ~~~~~ Antonio ~~~~ berada ~~~~ di ~~~~ negara ~~~~~ bagian ~~~~ apa ~~~~~ ?}}; 
		\node[word](ew0) [below left= of sent,xshift=28.2mm,yshift=4mm] {({\em San})}; 
		\node[word](ew1) [right = of ew0, xshift=-1mm] {({\em Antonio})};
		\node[word](ew2) [right = of ew1, xshift=-2mm] {({\em located})};
		\node[word](ew3) [right = of ew2, xshift=-2.5mm] {({\em in})};
		\node[word](ew4) [right = of ew3, xshift=-1.9mm] {({\em ~~~~~~~~~~~state~~~~~~~~~~~})};
		\node[word](ew6) [right = of ew4, xshift=-2.6mm] {({\em what})}; 
		\node[word](ew7) [right = of ew6, xshift=-2mm] {({\em ?})}; 
		
		\node[word](correctsem) [below = of sent, yshift=-2mm] {Gold Meaning Representation: $answer(loc(cityid('san\;antonio')))$};
		
		\node[word](htname) [below = of correctsem, xshift=-20mm, yshift=-1mm] {{\em Relaxed Hybrid Tree}};
		
		\node[word](hm1) [below = of htname, xshift=0mm] {$m_1$};
		\node[word](hw1) [below left = of hm1, xshift=0mm, yshift=-3mm] {{\em San Antonio breada di}};
		\node[word](hw5) [below right = of hm1, xshift=9mm, yshift=-3mm] {{\em ?}};
		\node[word](hm2) [below= of hm1, xshift=0mm, yshift=-3mm] {$m_4$};
		
		\node[word](hw34) [below= of hm2, xshift=0mm, yshift=-3mm] {{\em negara bagian apa}};
		
		
		
		\draw [chainLine] (hm1) to [] node[] {} (hw1);
		\draw [chainLine] (hm1) to [] node[] {} (hm2);
		\draw [chainLine] (hm1) to [] node[] {} (hw5);
		
		\draw [chainLine] (hm2) to [] node[] {} (hw34);
		%
		%

		
		\node[word](m1)[below= of htname, xshift=55mm, yshift=2mm] {$m_1$: Q{\small{UERY}} : $answer$ (S{\small{TATE}})};
		\node[word](m2)[below= of m1, yshift=4mm, xshift=-5.7mm] {$m_2$: S{\small{TATE}} : $loc$ (C{\small{ITY}})};
		\node[word](m3)[below= of m2, yshift=4mm, xshift=6.4mm] {$m_3$: C{\small{ITY}} : $cityid$ (C{\small{ITY}}N{\small{AME}})};
		\node[word](m4)[below= of m3, yshift=4mm, xshift=-6.4mm] {$m_4$: S{\small{TATE}} : $state$ ($all$)};
		\node[word](m5)[below= of m4, yshift=4mm, xshift=8.3mm] {$m_5$: C{\small{ITY}}N{\small{AME}} : ($'san\;antonio'$)};
		
		
		
		\node[word](wroot) [below = of sent, xshift = -55mm, yshift=-70mm] {{\em root}};
		\node[word](w0) [right = of wroot, xshift=2.5mm] {{\em San}};
		\node[word](w1) [right = of w0, xshift=1mm] {{\em Antonio}};
		\node[word](w2) [right = of w1, xshift=1mm] {{\em berada}};
		\node[word](w3) [right = of w2, xshift=1mm] {{\em di}};
		\node[word](w4) [right = of w3, xshift=1mm, yshift=-1mm] {{\em negara}};
		\node[word](w5) [right = of w4, xshift=1mm] {{\em bagian}};
		\node[word](w6) [right = of w5, xshift=1mm] {{\em apa}};
		\node[word](w7) [right = of w6, xshift=1mm] {{\em ?}};
		\node[word](dhtname) [below = of sent, yshift=-48mm] {{\em Dependency-based Hybrid Tree}};
		\draw [line width=0.8pt,->, color=fontgrey]  (wroot) to [out=48,in=132, looseness=0.9] node[above, yshift=-1mm, color=black]{$m_1$} (w3);
		\draw [line width=0.8pt,->, color=fontgrey]  (w3) to [out=140,in=40, looseness=1] node[above, yshift=-1mm, xshift=-1mm, color=black]{$m_2$} (w2);
		\draw [line width=0.8pt,->, color=fontgrey]  (w2) to [out=140,in=40, looseness=1] node[above, color=black, xshift= 0mm, yshift=-1mm]{$m_3$} (w1);
		\draw [line width=0.8pt,->, color=fontgrey]  (w1) to [out=120,in=60, looseness=1] node[above, yshift=-1mm, color=black, xshift=0mm]{$m_5$} (w0);
		\end{tikzpicture} 
	}
	\caption{Example results from \textsc{DepHT} and \textsc{RHT} on Indonesian.}
	\label{fig:indoerror}
\end{figure}
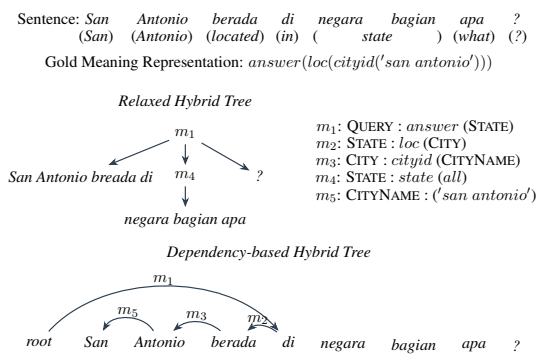

\subsection{Results and Discussion}

Table \ref{tab:nonneuralresults} (top) shows the results of our {dependency-based hybrid tree} model compared with non-neural models which achieve state-of-the-art performance on the GeoQuery dataset. 
Our model \textsc{DepHT} achieves competitive performance and outperforms the previous best system \textsc{RHT} on 6 languages.
Improvements on the Indonesian dataset are particularly striking (+11.8 absolute points in $F_1$).
We further investigated the outputs from both systems on Indonesian by doing error analysis.
We found 40 instances that are incorrectly predicted by \textsc{RHT} are correctly predicted by \textsc{DepHT}. 
We found that 77.5\% of the errors are due to incorrect alignment between words and semantic units.
Figure \ref{fig:indoerror} shows an example of such errors where the relaxed hybrid tree fails to capture the correct alignment.
We can see the question is asking ``{\em What state is San Antonio located in?}''. 
However, the natural language word order in Indonesian is different from English, where the phrase ``{\em berada di}'' that corresponds to $m_2$ (i.e., $loc$) appears between ``{\em San Antonio}'' (which corresponds to $m_5$ -- $'san\ antonio'$) and ``{\em what}'' (which corresponds to $m_1$ -- $answer$).
Such a structural non isomorphism issue between the sentence and the semantic tree  makes the relaxed hybrid tree parser unable to produce a joint representation with valid word-semantics alignment.
This issue makes the \textsc{RHT} model unable to predict the semantic unit $m_2$ (i.e., $loc$) as \textsc{RHT} has to align the words ``{\em San Antonio}'' which should be aligned to $m_5$ before aligning ``{\em berada di}''. 
However, $m_5$ has arity 0 and cannot have $m_2$ as its child. 
Thus, it would be impossible for the RHT model to predict such a meaning representation as output.
In contrast, we can see that our dependency-based hybrid tree representation appears to be more flexible in handling such cases. 
The dependency  between the two words ``{\em di}'' ({\em in}) and ``{\em berada}'' ({\em located}) is also well captured by the arc between them that is labeled with $m_2$.
The error analysis reveals the flexibility of our joint representation in different languages in terms of the word ordering, indicating that the novel dependency-based joint representation is more robust and suffers less from language-specific characteristics associated with the data.

\paragraph{Effectiveness of dependency}
To investigate the helpfulness of the features defined over latent dependencies, we conduct  ablation tests by removing the dependency-related features. 
Table \ref{tab:ablation} shows the performance of augmenting different dependency features in our \textsc{DepHT} model with basic features. 
Specifically, we investigate the performance of head word and modifier word features (\textsc{hm}) and also the bag-of-words features (\textsc{bow}) that can be extracted based on dependencies. 
It can be observed that dependency features  associated with the words are crucial for all languages, especially the \textsc{bow} features. 
\begin{table}[t!]
	\centering
	\setlength{\tabcolsep}{3pt} 
	\renewcommand{\arraystretch}{1} 
	\scalebox{0.72}{
		\begin{tabular}{|l|cccccccc|}
			\hline 
			& en & th&de&el&zh&id&sv&fa\\\hline 
			\textsc{DepHT} basic & 75.0 &82.1 & 70.4&74.6&76.1&71.9&73.9&69.3\\ 
			\textsc{basic}+\textsc{hm} feats. & 80.7 & {\bf 83.9} & {\bf 75.7} &79.2&81.1&85.0&81.1&72.5\\
			\textsc{basic}+\textsc{bow} feats. & {\bf 86.1}&83.2&73.9&{\bf 79.3}&{\bf 81.4}& {\bf 86.1} &{\bf 85.4}&{\bf 73.2} \\\hline
			\textsc{DepHT} & 86.8& 81.8& 76.1 & 80.4& 81.4& 86.8& 85.4& 73.9 \\\hline
		\end{tabular}
	}
	\vspace*{-2mm}
	\caption{$F_1$ scores of our model with different dependency features.}
	\label{tab:ablation}
	\vspace*{-3mm}
\end{table}

\paragraph{Effectiveness of neural component}
The bottom part of Table \ref{tab:nonneuralresults} shows the performance comparison among models that involve neural networks.
Our \textsc{DepHT} model with embeddings as features can outperform neural baselines across several languages (i.e., Chinese, Indonesian and Swedish). 
From the table, we can see the neural component is effective, which consistently gives better results than \textsc{DepHT} and the approach that uses word embedding features only.
\citet{susanto2017semantic} presented the \textsc{Neural HT} model with different window size $J$ for their multilayer perceptron. Their performance will differ with different window sizes, which need to be tuned for each language. 
In our neural component, we do not require such a language-specific hyperparameter, yet our neural approach consistently achieves the highest performance on 7 out of 8 languages compared with all previous approaches.
%
As both the embeddings and the neural component are defined on the dependency arcs, the superior results also reveal the effectiveness of our dependency-based hybrid tree representation. 


\section{Conclusions and Future Work}

In this work, we present a novel {\em dependency-based hybrid tree} model for semantic parsing. 
The  model captures the underlying semantic information of a sentence as latent dependencies between the natural language words. 
We develop an efficient algorithm for exact inference  based on dynamic-programming.
Extensive experiments on benchmark dataset across 8 different languages demonstrate the effectiveness of our newly proposed representation for semantic parsing.

Future work includes exploring alternative approaches such as transition-based methods
~\cite{nivre2006maltparser,chen2014fast} for semantic parsing with latent dependencies, 
applying our dependency-based hybrid trees on other types of logical representations (e.g., lambda calculus expressions and SQL~\cite{P18-1033}) as well as multilingual semantic parsing~\cite{jie2014multilingual,susanto2017neural}.

%

\section*{Acknowledgments}
We would like to thank the anonymous reviewers for their constructive comments on this work. 
We would also like to thank Yanyan Zou for helping us with running the experiments for baseline systems.
This work is supported by Singapore Ministry of Education Academic Research Fund (AcRF) Tier 2 Project MOE2017-T2-1-156, and is partially supported by project 61472191 under the National Natural Science Foundation of China.

\bibliography{dht}
\bibliographystyle{dht}

\end{document}